**Does GPT-4 surpass human performance in linguistic pragmatics?**

Ljubiša Bojić[1,2,3], Ph. D., Senior Research Fellow
(Corresponding author; Email address: ljubisa.bojic@ivi.ac.rs; ORCID: 0000-0002-5371-7975)

Predrag Kovačević[4], Ph. D., Assistant Professor
(Email address: predrag.kovacevic@ff.uns.ac.rs; ORCID: 0000-0003-2545-9240)

Milan Čabarkapa[5], Ph. D., Assistant Professor
(Email address: mcabarkapa@kg.ac.rs; ORCID: 0000-0002-2094-9649)

[1] Institute for Artificial Intelligence Research and Development of Serbia; Address of correspondence: 1 Fruskogorska, Novi Sad, Serbia;
[2] University of Belgrade, Institute for Philosophy and Social Theory, Digital Society Lab; Address of correspondence: 45 Kraljice Natalije, 11000 Belgrade, Serbia;
[3] Complexity Science Hub; Address of correspondence: Metternichgasse 8, 1030 Vienna, Austria;
[4] University of Novi Sad, Faculty of Philosophy; Address of correspondence: 2 Dr Zorana Djindjica, 21102 Novi Sad, Serbia;
[5] University of Kragujevac, Faculty of Engineering; Address of correspondence: 6 Sestre Janjic, 34000 Kragujevac, Serbia;

## Abstract

As Large Language Models (LLMs) become increasingly integrated into everyday life as general-purpose multimodal AI systems, their capabilities to simulate human understanding are under examination. This study investigates LLMs' ability to interpret linguistic pragmatics, which involves context and implied meanings. Using Grice's communication principles, we evaluated both LLMs (GPT-2, GPT-3, GPT-3.5, GPT-4, and Bard) and human subjects (N = 147) on dialogue-based tasks. Human participants included 71 primarily Serbian students and 76 native English speakers from the United States. Findings revealed that LLMs, particularly GPT-4, outperformed humans. GPT-4 achieved the highest score of 4.80, surpassing the best human score of 4.55. Other LLMs performed well: GPT-3.5 scored 4.10, Bard 3.75, and GPT-3 3.25; GPT-2 had the lowest score of 1.05. The average LLM score was 3.39, exceeding the human cohorts' averages of 2.80 (Serbian students) and 2.34 (U.S. participants). In the ranking of all 155 subjects (including LLMs and humans), GPT-4 secured the top position, while the best human ranked second. These results highlight significant progress in LLMs' ability to simulate understanding of linguistic pragmatics. Future studies should confirm these findings with more dialogue-based tasks and diverse participants. This research has important implications for advancing general-purpose AI models in various communication-centered tasks, including potential application in humanoid robots in the future.

## Introduction

The proliferation of artificial intelligence (AI) and Large Language Models (LLMs) such as language translator systems, personal assistants, chatbots, and the like, is today more pervasive than ever. The progress in this field has significant social implications, ranging from automating tasks to aiding in various professional operations and reshaping how we communicate and interact with technology.

The evolution and integration of LLMs into various segments of society have sparked significant interest due to their human-like text generation capabilities. From customer service and education to healthcare and entertainment, these LLMs offer an array of functionalities that are reshaping our daily interactions and engagements (Stefanowicz, 2022).

As these models permeate society, the extent of their capabilities and their potential impact on societal structures continue to be topics of robust research and discussion (Thunström, 2022; Zhang & Li, 2021; Alipour et al., 2024; Bojic, 2022; Bojic, 2024; Bojic et al., 2024a, 2024b). Recent studies have begun to explore the personality traits inherent in LLMs (Bodroza et al., 2023; Li et al., 2022; Kumar et al., 2022; Skjuve et al., 2022) and alignment with certain social values, ideologies, moral foundations (King, 2022; Wu et al., 2023; Rutinowski et al., 2023; Hartmann et al., 2023; Simmons, 2022; Abdulhai et al., 2023; McGee. 2023).

With advances in artificial intelligence, these models have begun simulating cognitive abilities that mimic human-like traits – a significance that extends far beyond their interactive capabilities. For instance, they exhibit a sense of perception and expression, reasoning abilities, and spontaneity, leading to an increasingly complex imitation of human-like characteristics (Binz & Schulz, 2022; Kosinski, 2023). The ability of these models to assume a persona and display traits and capabilities, such as theory of mind, directly influences the perception and acceptance of these models by society (Kosinski, 2023).

Recent rapid developments in LLMs, particularly OpenAI's GPT series and Google's Bard, have heightened their capabilities, pushing them close to human-level linguistic and cognitive performance (Radford et al., 2019). These models claim to have an enriched understanding of semantics, syntax, and context, making them more adept at comprehending and generating text that is indistinguishable from human-produced text (Brown et al., 2020).

Various studies have evaluated these models based on their responses to diverse linguistic tasks, and the results have shown these models to be increasingly good at emulating human-level understanding (Radford et al., 2019; Gao et al., 2024; Amirizaniani et al., 2024; Barnard & Otte, 2024; Xie & Zou, 2024). However, most of these evaluations have largely focused on the output quality and fluency but have largely ignored the aspect of linguistic pragmatics.

### *Linguistic pragmatics*

Linguistic pragmatics represents an essential aspect of communication studies and is an established field within the realm of linguistics. Originally conceptualized by analytic philosophers John Langshaw Austin and Paul Grice in the 1950s, pragmatics refers to the subfield of linguistics that studies how context influences the interpretation of meaning (Grice, 1975). Fundamentally, it concerns the way in which language is used in social interactions, emphasizing how context, world knowledge, and the speaker's goal can all influence the actual meaning of a phrase or sentence.

This extends well beyond the words' literal interpretation to incorporate various contextual elements necessary for inferring the speaker's intended meaning.

Existing research within this domain underlines the richness and multidimensionality of pragmatics, with scholars highlighting its importance in varying contexts right from understanding cultural aspects, facilitating sociolinguistic interactions, to enhancing machine learning algorithms (Levinson, 1983; Thomas, 1995; Huang, 2006; Yule, 2006). Levinson (1983), for instance, posits that understanding pragmatic phenomena is paramount to appreciating how human language functions as a communication system. The ability to utilize and comprehend language in a pragmatic sense—understanding sarcasm or irony, inferring unspoken conclusions, detecting deception, and more— is crucial for successful human-to-human interaction.

Thomas (1995) calls attention to the fact that cultural variations often lead to different pragmatic rules, contributing to miscommunication instances during cross-cultural communication. By studying pragmatics, we gain a better understanding of these differences, thereby minimizing such miscommunications. Indeed, for any effective communicative process, understanding the conventionally agreed-upon meanings of words or grammatical rules is necessary but not sufficient. The real challenge lies in grasping meaning beyond the surface, considering elements such as inference, implicature, presupposition, and speech acts that form the crux of pragmatic communication.

In the realm of artificial intelligence (AI), a pragmatic understanding of language can be instrumental in improving machine learning algorithms' language comprehension abilities (McCarthy & Hayes, 1981; Minsky, 1981; Russell & Norvig, 2010). For AI systems to interact in a manner similar to human interlocutors, they must simulate that language is often used to perform action, to engage in social interaction, and to convey more than what is being explicitly stated. Hence, the concept of pragmatics and its various theories, such as speech act theory (Austin, 1962; Searle, 1969), politeness theory (Brown & Levinson, 1987), and relevance theory (Sperber & Wilson, 1995), become eminently relevant in the context of AI and machine learning.

Pragmatic competence is critical in diverse capacities, influencing interpersonal relationships, socio-cultural interactions, professional communication, second language acquisition, and language processing in AI systems. It offers a lens to understand the complexities, intricacies, and implicit rules of human communication. As such, pragmatic competence is a valuable and practical skill to acquire and an important area of exploration for any model that aims to interpret human language effectively.

### *Hypothesis*

This study aims to assess the ability of LLMs in comprehending and interpreting linguistic pragmatics, drawing upon Grice's communication principles and using various dialogue-based tasks. Therefore, the following hypothesis will guide this examination:

H1: Humans will outperform LLMs when it comes to correctly interpreting the pragmatic meaning of dialogues.

This study looks to fill an essential gap in the research on LLMs, showcasing their ability and limitations in the realm of linguistic pragmatics. The findings will have broad implications for the practical applications of LLMs and future development in AI language understanding.

## Methodology

This section presents the research framework used in this study. The explanation includes the theoretical underpinnings of the concepts involved, an overview of the distinct models employed, and a clarification on how participants were chosen. The section, segmented into parts such as tests, task scope, participants in the study, AI models, grading of responses, method of triangulation and data and reproducibility, further explains the specialized techniques used for assessing participant responses.

### *Tests*

The most suitable theoretical framework for this type of investigation would be Sperber and Wilson's (1986, 1995) Relevance Theory building on the work of Paul Grice and his Theory of Implicature (Grice, 1989). Grice is concerned with the rules or principles that govern conversations/communication. The main principle he identifies is the Cooperative Principle, according to which the participants in a conversation are expected to cooperate or work together to keep the conversation going and allow both of them to benefit (learn something, have fun, etc.) from it (Grice, 1975). The Cooperative Principle is broken down into four maxims:

1. Maxim of Quantity: provide the optimal amount of information in your statements (do not „beat around the bush" and do not make your statements too dense with information)
2. Maxim of Quality: do not make assertions that you know to be false
3. Maxim of Relation: do not make assertions that provide information that has no bearing on the previous discourse (be relevant)
4. Maxim of Manner: be clear, avoid ambiguity and obscure expressions, etc.

Grice's work on implicatures stems directly from the formulation of the Cooperative Principle because he observes that in some cases sentences that individuals produce appear to violate or flout one or more of the maxims he identified, and yet the conversation does not break down. What is more, the very act of flouting the maxims is what allows a particular message to be conveyed. This is where implied meanings or implicatures enter the picture, i. e. in some conversations, the flouting of one or more of Gricean maxims is the signal that the sentence that is uttered is NOT to be interpreted literally. Instead, it gives rise to conversational implicatures, or non-literal (i.e. implied) meanings that are recoverable from the context if the literal meaning of the statement is taken as only one of the pieces of evidence needed to interpret the message along with non-verbal clues, cultural knowledge, the setting in which the situation takes place, etc. Consider the example in (1).

(1) Two men are standing at a bus stop in pouring rain. One of them says, "the weather is absolutely marvelous today", and the other one responds, "it simply can't get any better".

In (1), the first speaker is obviously flouting the Maxim of Quality by making a statement that is obviously false. However, the second speaker, does not contradict him because he understands that the utterance was not intended to convey its literal meaning but a sense of irony. For this reason, he responds, in a similar fashion. In that sense, the conversation is directed towards entertainment rather than informativeness.

In addition to conversational implicatures (i.e. implied meanings that are derived from the general context in which the sentence is uttered), Grice also identifies conventional implicatures. These are implied meanings that are triggered by certain words or phrases independently of the context in which they are used. For example, the conjunction “but“ is taken to imply a contrast between the notions that it connects as in (2).

(2) Peter is handsome but smart.

If one utters something like (2), the message is that the quality of being handsome stands in some kind of contrast with the quality of being smart (i.e. the combination of these two is unlikely). This implication will be present in every use of this conjunction irrespective of the context.

Let us illustrate these notions on several simple examples. Consider first the situation described in (3).

(3) A 40-year old man invited a 25-year old woman to dinner. They have a lively conversation. At the end, the man asks the woman: do you want to come up to my apartment to see my collection of butterflies. The woman responds: “Sorry, I have a boyfriend”.

In (3), the woman’s response obviously flouts the Maxim of Relation because the literal meaning of the response does not seem to be relevant in any way in the context of the man’s invitation. However, if the man’s invitation is also interpreted in a non-literal fashion, namely, as an invitation for a romantic encounter, then, the literal meaning of the response becomes fully relevant.

Next, consider the situation in (4)

(4) A young man approaches a person at a party and says “Hello, my name is Stefano”. The person replies: “Mine isn’t”.

In the example situation given in (4), if the young man says „My name is Stefano“ in a friendly manner and the addressee response “mine isn’t“, there are several possible analyses of the situation. The addressee might not be interested in who Stefano is at all, in which case Stefano’s approach and introduction could be seen as a violation of the Maxim of Relation, which is subsequently signalized by the addressee in his response. Alternatively, the addressee might accept Stefano’s polite introduction as part of customary behavior at parties whereby one is expected to respond by introducing him or herself back. In this sense, the response “mine isn’t” could be seen as a deliberate flouting of the Maxim of Quantity (merely saying that one’s name is not Stefano is not informative enough as the addressee probably wants to know what one’s name *is*).

Finally, consider the example in (5).

(5) Two academics are chatting over lunch. One of them asks the other about how many people had attended a guest lecture she delivered the previous week. She responds, “some of the faculty attended”.

The response in (5) derives a special communicative effect not from conversational but from conventional implicature licensed by word choice similar to example (2).. While the literal meaning of the statement does not exclude the possibility that all faculty members as well as (a large number of) students attended the talk, by choosing the word „some“ (and not „all“) and mentioning only the faculty and not the students, the speaker implies that the talk was attended only by the faculty, and, worse, that only a portion of the faculty members came to the talk. The overarching implication is that the talk was very poorly attended.

As we already suggested and as the examples (2-5) illustrate, the existence of Gricean Maxims shapes human communication in an important way and the possibility of flouting them only serves to highlight their importance as evidenced by the rather specific communicative effect that it achieves. In that light, to achieve complete pragmatic or communicative competence, AI must be capable of not only abiding by these maxims but also flouting them and recognizing the (potential) communicative effects that such linguistic maneuvers might achieve. However, one must recognize that the degree of pragmatic competence can show a high degree of variance among native speakers of a particular language (cf. Hymes' 1972 classic remarks on what he terms 'Communicative Competence'), which is why the assessment of the level of pragmatic competence exhibited by a particular AI model can only be carried out based on an appropriate benchmark of the average or expected human pragmatic competence.

To obtain empirical data that would allow for a comparison of pragmatic competence exhibited by different AI models as well as human subjects, a series of 20 separate conversation examples were devised (Appendix 1), each illustrating a violation of one of the Gricean maxims. This set includes four unique instances for each of the four maxims and an additional four exemplifying conventional implicature. These examples noted further in this text were constructed to serve as viable testing material for the study.

The reason for using only 20 conversations lies largely in the consideration towards human participants. Research methods involving human subjects usually aim to balance the need for comprehensive data collection with respect for the participants' time and cognitive load. In this study, each participant is required to engage with each conversation and subsequently answer a survey or a series of questions about their impression of the conversation. This process requires significant mental effort and time investment from each participant.

To continue with a larger number of conversations could potentially lead to fatigue in human participants, skewing data accuracy and compromising the reliability and validity of participant responses. Thus, although more conversations could potentially provide richer data, it is impractical and unethical to overburden human subjects in this manner. The choice of twenty conversations represents a compromise between the data collection needs of the research and the practical limitations of human participation.

This study aims to provide meaningful conclusions about the pragmatics of conversation in AI, while also respecting ethical principles involving human subjects' participation in research. Therefore, the restriction to 20 conversations is both a practical and ethical precaution, ensuring valid and reliable data without causing undue burden to human participants.

Based on this, dialogues used in research are formed and presented in the Appendix 1.

While the dialogues in this study were constructed rather than drawn from authentic conversational corpora, this approach was chosen to ensure consistent illustration of specific pragmatic phenomena (i.e., flouting of particular Gricean maxims). By designing each example to target a single pragmatics concept, we aimed to systematically probe the capabilities of both LLMs and human participants in identifying these targeted violations.

Additionally, LLMs may have already 'assessed' existing corpora, given that they are trained on vast amounts of text. Therefore, using original, fabricated examples is essential to prevent any advantages AI systems might have over human participants.

Nonetheless, we acknowledge that the use of constructed dialogues has limitations. In real-world communication, factors such as spontaneous topic shifts, overlapping speech, interruptions, and other contextual elements may influence conversational flow in ways our constructed vignettes do not fully capture. Additionally, it was impossible to find such examples in existing corpora because corpora are generally not annotated for global pragmatic effects like sarcasm, thereby hindering the selection of spontaneously occurring dialogues that capture such nuanced aspects of communication.

*Task Scope*

While each of the twenty dialogues used in this study was selected because it exemplifies one or more violations of Gricean maxims, the prompt presented to participants and models—"How do you interpret this dialogue?"—is not strictly or exclusively pragmatic. Instead, it is an open-ended question designed to capture their overall interpretation, spanning both literal and implied meanings.

Our grading framework accordingly uses a continuum ranging from "poor understanding of the literal meaning embedded in the dialogue" (score of 1) to "clear understanding of the core aspect of the dialogue" (score of 5). This general approach ensures that interpreters who completely misunderstand the scenario (e.g., misidentifying the speaker's statements or the basic facts of the situation) are distinguished from those who not only grasp the surface-level content but also recognize and address any deeper communicative challenges—whether those challenges stem from sarcasm, irrelevance, ambiguous language, or some other aspect of dialogue that commonly falls under linguistic pragmatics.

In other words, while our dialogue set is built around Gricean violations to illustrate typical pragmatic phenomena (e.g., sarcasm, irony, irrelevance, ambiguity), the interpretive task itself is broad: participants and models are free to focus on literal aspects, contextual keys, or inferred meaning. Those demonstrating deeper insight into how the dialogue "works", particularly when the intended meaning contradicts the literal meaning, earn higher scores on the five-point scale. Hence, although pragmatics is central to our data selection, the evaluation extends to general language comprehension, balancing literal understanding with the ability to detect and explain any pragmatic implications.

*Participants in the study*

All participants were informed that the experiment examined how individuals interpret dialogues (see "Informed Consent"). However, they were never explicitly told that the study's main focus was on pragmatic phenomena or Gricean violations. Our goal was to capture each participant's natural interpretive strategies without making them "hunt" for specific linguistic maxims or forms of irony. Nonetheless, because the dialogues were presented sequentially, some participants may have noticed recurring patterns—such as sarcasm or misdirection—and thus gained insights into the nature of the task. This possibility was especially pronounced among participants with any linguistic background or enhanced sensitivity to pragmatic cues. Their responses occasionally reveal an awareness that the dialogues might contain deliberate violations of conversational norms.

**First Cohort.** In an initial research, human participants were non-native speakers of English. They were enlisted through an advertisement disseminated across the campus and examined in person on the 12th of September, 2023, within the confines of the Faculty of Engineering Sciences. Given that Serbian high school graduates are expected to attain at least a B2 (upper-intermediate) CEFR level by the end of secondary education, and that many university students progress toward a C1 (advanced) standard, we considered advanced English competence crucial for a fair assessment of the research tasks. To ensure an adequate command of English for the task, each individual was initially screened using a placement test that assessed vocabulary, grammar, and reading comprehension skills. Only those scoring at an advanced (CEFR C1–C2) were invited to participate. Although non-native speakers can sometimes underperform relative to native speakers in linguistic tasks (due to lexical gaps, interference from their first language, or other factors), the specific design of this study minimizes many of those concerns.

First, the core challenge—interpreting potential communicative misalignments or conversational "traps"—does not hinge primarily on specialized or idiomatic usage; rather, it focuses on understanding context, intent, and basic conversational implications. By selecting participants with demonstrated competence in English, we confirmed they had sufficient grammatical and lexical knowledge to grasp and interpret the dialogues' literal meanings. Second, in our grading process, responses were evaluated on whether they identified the main interpretive issues (e.g., sarcasm, irrelevance), not on grammatical or stylistic quality. Minor language inaccuracies did not adversely affect scoring. Thus, the risk that limited proficiency would confound the core interpretive element of the task is reduced.

Out of 110 applied participants in the first Serbian Cohort, 71 were selected after a pre-test for the English language was administered to them on September 10, 2023. The selected cohort (N=71) was predominantly comprised of Caucasian respondents (82%), with the majority (93%) coming from Serbia. The gender breakdown showed a higher proportion of males (61%) than females (38%).

A total of 71 individuals took part in the survey. Their ages ranged from 18 to 54 years, with the largest proportion (n=43) falling in the 18–24 age group, followed by those aged 35–44 (n=16), 25–34 (n=9), and 45–54 (n=3). Regarding marital status, the majority (n=56) reported being single or never married, while 11 participants indicated they were married or in a domestic partnership, 1 participant was divorced, and 3 declined to specify their marital status.

Regarding employment status, nearly half the participants (47%) identified themselves as students. Approximately one-third (32%) reported full-time employment, while 10% were unemployed and looking for work, and 9% held part-time jobs. A small proportion was self-employed (3%), and 1% preferred not to disclose their employment status.

**Second Cohort.** In addition to an initial Serbian-based sample, a second cohort of participants was recruited through the Latenta Market Research Agency on January 29, 2025 (Latenta, 2025). A total of 567 responses were received, of which 150 were complete, 64 partially completed, and 353 dropped out. Following further screening by three native English-speaking evaluators, 73 participants were disqualified due to failing attention checks or providing inconsistent responses. The 76 individuals with complete responses formed our final dataset for the second cohort.

All 76 respondents self-identified as native speakers of English, and every participant both resided in, and originated from, the United States. The majority identified as White (84.21%), followed by Black or African American (7.89%) and Hispanic, Latino, or Spanish origin (5.26%). An additional 2.63% preferred not to disclose their ethnicity.

In terms of age, 5.26% of participants were between 18 and 24 years old, 15.79% were 25–34, 13.16% were 35–44, 10.53% were 45–54, 21.05% were 55–64, and 34.21% were aged 65 or older. Women comprised 59.21% of the sample, and men accounted for 40.79%. A plurality (43.42%) were married or in a domestic partnership, followed by single/never married (32.89%), divorced (15.79%), widowed (5.26%), and separated (1.32%), while 1.32% chose not to respond.

Educational backgrounds ranged from 21.05% having a high school diploma or equivalent, 3.95% reporting trade/technical/vocational training, 19.74% having some college credit, and 14.47% holding an associate degree. About a quarter (26.32%) had earned a bachelor's degree, while 10.53% reported a master's degree, 1.32% a doctoral degree, and 2.63% a professional degree.

Regarding employment, 28.95% were employed full-time, 9.21% part-time, and 35.53% were retired. Meanwhile, 5.26% identified as homemakers or caregivers, 3.95% reported self-employment, 10.53% were not employed but searching for work, 2.63% were not employed and not seeking work, and 3.95% were unable to work due to disability or health conditions.

The second cohort provided a broad cross-section of native English speakers that complemented the primarily non-native speaker profile of our first cohort. This design enabled comparisons across varied linguistic and cultural backgrounds while ensuring that sufficient English language proficiency and relevant demographic diversity were represented in the data.

After evaluating the second cohort, English-speaking graders also assessed the Serbian sample of responses. Their evaluations confirmed the scores initially given by the primary graders of that sample, thereby making the two groups comparable.

*AI Models*

The objective of this study was to explore the field of linguistic pragmatics by utilizing an assortment of models provided by OpenAI and Google's Bard. The rationale for selecting these specific models stemmed from their respective stages of development and their inherent capabilities. The study was carried out by utilizing the OpenAI Playground, employing GPT 2, GPT 3, GPT 3.5, and GPT 4 large language models, on the 30th of August 2023. The settings of the models were default during the testing.

Subsequent to this, Google's Bard platform was used to execute the study on the 31st of August 2023. The responses gathered from this research have been made publicly accessible (OSF, 2023). To ensure reliability, the testing procedure for the Large Language Models (LLMs) was repeated on December 9, 2023, confirming confidence in the stability and consistency of their performance.

To parallel how humans encountered one dialogue at a time, each Large Language Model was also presented with individual dialogues followed by the question, “How would you interpret this dialogue?” Although each question was asked separatly, not as a batch, the session context (i.e., the conversation history with the model) was not reset between question–dialogue pairs. As a result, the models could draw upon prior interactions and potentially recognize patterns or glean that multiple dialogues involved conversational “traps.” Although these older models (GPT2, GPT3) did not appear to benefit much from this incremental context—likely due to their limited instruction tuning and weaker capacity for cross-turn reasoning—newer systems (GPT3.5, GPT4, Bard) often gave progressively more refined answers. Similarly, over time, human participants could pick up on recurring themes in the dialogues, even if they had not been explicitly told these were examples of Gricean flouting.

*Grading of responses*

The methodology for examining linguistic pragmatics within this study unfolded through a rigorous and detailed process. This process began by presenting a narrative to each of the OpenAI and Bard models, eliciting their interpretations by asking the question, "How would you interpret this dialogue?"

As they responded, their responses were scrutinized and evaluated on a scale of 1 to 5 based on set judgement criteria.

The scale was constructed as follows: a score of '1' implied that a model's response reflected a poor simulation of understanding of the literal meaning embedded within the dialogue. This misunderstanding would be displayed through an inconsistent retelling of the narrative or a misguided emphasis on irrelevant aspects and details.

A middle-ground score of '3' specified that while a model had demonstrated a good simulation of understanding of the dialogue's literal meaning, it fell short in addressing the broader implications or the reasoning behind the encountered communicative challenges, stemming from the flouting or violation of the dialogue's contextual maxim. Alternatively, the focus might have erroneously been on non-core issues.

The highest score, '5', was reserved for the model's response that successfully illustrated a clear simulation of understanding of the core aspect of the dialogue - the communicative challenges, born out of the deliberate violation or disregard for the dialogue-specific maxims.

The grading system was accompanied by a penalty system. A deduction of 1 to 2 points was enforced when the model, despite correctly simulating identification of the problem, dove into unnecessary speculative additions (refer Bard's response to question No. 5 – OSF (2023)), or wrongfully identified the party in the dialogue that caused the communicative challenge by violating or flouting the maxim (refer GPT 3's response to question No. 11 and juxtapose it with the responses of the later versions; Bard exhibits a similar behavior as GPT 3).

*Method of triangulation*

The process of interpretation grading was conducted by the co-authors through a rigorous method of triangulation. The first and second authors evaluated the interpretations individually. Any discrepancies that arose from their separate evaluations were then resolved through the insightful arbitration of the third author. With this structure, the study adopted a balanced, flexible, yet rigorous approach to evaluating the models' performances on linguistic pragmatics.

All written interpretations from both human participants and LLMs were compiled and initially evaluated by two co-authors (the first and second authors) independently. Each expert applied the same five-point grading scale (outlined above) to gauge the extent to which the response recognized the communicative challenges posed by the broken maxim(s).

Once these initial evaluations were complete, the two sets of scores were compared, and any discrepancies were flagged for further review. For example, in one instance involving Dialogue 5 (flouting the Maxim of Quality), the first author assigned a score of "4," emphasizing that the interpretation captured the literal meaning and partially addressed the sarcasm by referencing the absurdly positive description of a decrepit apartment. The second author assigned a "5," arguing that the interpretation explicitly highlighted the sarcastic contrast between what was said ("the most luxurious suite") and the actual deplorable condition of the apartment. When such a discrepancy arose, the third author acted as an arbiter, reviewing both evaluations in the context of the dialogue.

In the arbitration phase of our study, we undertook a systematic approach to ensure the integrity of evaluative judgments. First, we retraced the dialogue's intended pragmatic violation or flouting, thereby establishing a foundational understanding of the communicative intent. This step was essential for contextualizing subsequent evaluations.

Next, we revisited each evaluator's reasoning against our predefined grading criteria. This reassessment enabled us to detect any deviations or inconsistencies in judgment, ensuring that all evaluations were aligned with the established standards.

Finally, we considered potential oversights that could affect the accuracy of evaluations. For example, we examined instances where one expert might have disproportionately emphasized gestural or contextual cues not present in the written text, or alternatively, where undue weight was assigned to extraneous details. By addressing these factors, our arbitration process aimed to reconcile divergent interpretations and fortify the reliability of the evaluative framework.

In the above example, the third author concluded that the explicit demonstration of sarcastic contrast was indeed central to the clip, and thus aligned with the higher rating. Consequently, the final expert score for that interpretation was set at "5." In other cases, the discrepancy might be resolved by averaging scores, or if both authors misunderstood a crucial contextual clue, the final rating could shift substantially (e.g., from 5 to 3, or vice versa).

This arbitration step ensures that the final rating reflects a broad consensus on the interpretation's quality. This systematic approach minimized individual bias and clarified, for each dialogue, which aspect of the communicative challenge (e.g., irony, irrelevance, or misdirection) was being recognized—or overlooked—by the model or participant under review.

### *Data and reproducibility*

During the entirety of the prompting processes, all data are securely and permanently saved on LLM platforms with links noted in OSF (2023). This guarantees the availability of all collected information throughout and beyond the span of the research project. This establishes a comprehensive record of all interactions, while also providing a basis for transparent evaluation and reproducibility of the research findings.

All the gradings, along with hyperlinks to primary data on both OpenAI and Bard platforms, are made publicly accessible through the Open Science Framework (OSF, 2023). This ensures full transparency of the data, making it accessible for further investigation, validation, or replication.

The framework used for grading the conversations is clearly explained and detailed in Appendix 2. This offers an understanding of the evaluation criteria and grading system used for the study, enabling its application in similar future studies and allowing for a comparative analysis across different contexts. This transparency ensures the reliability and validity of the research and allows for its results to be critically appraised by the scholarly community.

## Results

The Results section presents the findings of the investigation in relation to the theoretical framework described previously, essentially the performance evaluation of Large Language Models (LLMs) versus two cohorts of human subjects in linguistic pragmatics based on the Gricean maxims. These results offer an insight into the progression of LLMs and the varying degrees of simulating pragmatic understanding across different models and human subjects. Table 1 offers an in-depth comparative analysis between the different participants, providing specifics on their average scores. LLMs showcased better performance in terms of average scoring, with GPT 4 emerging as the highest scorer.

Results are listed in Table 1. The table includes 147 human respondents from two cohorts (N=71 primarily Serbian students and N=76 native English speakers from the US) and 5 LLMs, all ranked by their average linguistic-pragmatics scores. Additionally, the table presents separate average scores for each human cohort and the LLM group, totaling 155 entries.

LLMs performed better than humans both on average (3.30 vs. 2.8) and individually (4.8 vs. 4.55), which rejects the hypothesis (H1).

Table 1. Overview of scores for human respondents, language learning models (LLMs), and their respective averages.

| No. | Subject (both humans and LLMs) | Average score |
|---|---|---|
| 1 | GPT 4 | 4.80 |
| 2 | Best Human - Score | 4.55 |
| 5 | GPT 3.5 | 4.10 |
| 10 | Bard | 3.75 |
| 18 | LLM Average Score | 3.39 |
| 23 | GPT 3 | 3.25 |
| 46 | Human Average Score 1 (First Cohort – Serbian students) | 2.80 |
| 93 | Human Average Score 2 (Second Cohort –US participants, native speakers of English) | 2.34 |
| 155 | GPT 2 – Lowest Score | 1.05 |

**Discussion**

*Evolution of AI Models in Linguistic Pragmatics*

The progression of GPT models when it comes to pragmatics can be illustrated based on responses to dialogue No. 16 - starting from GPT 2 which does not simulate understanding of the dialogue at all, over GPT3 and 3.5 which show some simulation of understanding of the issue at hand, and finally ending with GPT4 which directly simulates identifying “ambiguous language” as the source of the problem (essentially explaining the maxim at hand).

GPT3 gives very succinct responses which are to the point but lack details. Often, these brief responses do not explicate what the source of the communicative challenge might be at all and goes directly into interpretation of the implications of what is said (see interpretation of dialogue No. 9).

GPT3.5 gives a detailed description of the literal exchange in the dialogue followed by a comment on pragmatics (see the response to dialogue No. 7).

Bard tends to give lengthy responses, often evaluates the stylistic and rhetorical features of the dialogue (see the responses to dialogues No. 8 and 9).

Related to the same point, there is an amazing jump from GPT3.5 to GPT4 in response to dialogue No. 17.

GPT4 explains exactly what the point of the dialogue is referencing humorous play with mathematical logic as opposed to typical language use, this again explains the source of the problem directly.

The findings indeed reject the Hypotheses (H1), signifying the superior ability of LLMs in simulating linguistic pragmatics understanding when compared to the human subjects. On a note of progression, GPT 4 demonstrated an impressive leap in interpretation accuracy from its preceding versions, specifically in the case of dialogue 17. The results indicate a promising future for LLMs in linguistic pragmatics interpretation, while also offering areas of analysis not addressed as part of this study.

*Differences between the two cohorts*

The examination of linguistic pragmatics within this study has revealed differences in performance between human cohorts. Notably, the first cohort of Serbian students outperformed the second cohort of U.S. participants in average scores (2.80 vs. 2.34). Several key factors may explain the better performance of the Serbian cohort.

The Serbian participants were rigorously screened for advanced English proficiency, requiring a minimum of CEFR C1–C2 level. This ensured that all participants had a strong command of English, particularly in reading comprehension and language interpretation skills. In contrast, the U.S. participants were self-reported native English speakers without additional language proficiency screening. While native fluency suggests a high level of language proficiency, it does not guarantee advanced interpretative skills, especially in linguistic pragmatics.

The second cohort predominantly consisted of university students who were actively engaged in academic study, primarily in fields that may enhance critical thinking and analytical skills. Their educational environment likely placed a strong emphasis on language, literature, and communication skills, which are directly relevant to tasks involving linguistic pragmatics. On the other hand, the U.S. cohort displayed a broader range of educational backgrounds, with only about 26% holding a bachelor's degree and a significant portion having high school education or some college experience. This variation may have influenced their ability to interpret complex pragmatic cues.

The Serbian participants were mostly younger adults aged between 18 and 24 years, an age group that is often engaged in higher education and may be more accustomed to test-taking and cognitive tasks associated with academic assessments. The U.S. participants had a wider age range, with over half aged 55 and above. Cognitive processing speed and certain types of problem-solving abilities can vary with age, potentially impacting performance on tasks requiring quick interpretation of dialogues.

The Serbian students participated in the study within an academic setting, which may have heightened their motivation and engagement. They might have perceived the task as part of their educational activities, leading to a more diligent approach. Conversely, the U.S. participants were recruited through a market research agency, and their level of engagement may have varied. Participation through such platforms might not elicit the same degree of motivation, potentially affecting task performance.

The Serbian participants completed the tasks in person and potentially under similar conditions, which could contribute to consistency in their responses. The U.S. participants completed the survey remotely, leading to variability in testing environments that might introduce distractions or inconsistencies affecting performance.

These factors highlight the complexity of comparing human performances across diverse cohorts. The superior performance of the Serbian students does not necessarily indicate that non-native speakers are better at interpreting linguistic pragmatics than native speakers in general. Instead, it reflects the specific characteristics of the sample populations in this study.

The findings highlight the importance of considering participant demographics, language proficiency levels, and testing conditions when interpreting results in studies involving language comprehension and pragmatics. They suggest that with appropriate training and proficiency, non-native speakers can perform at levels comparable to or even exceeding those of native speakers in specific linguistic tasks.

## Conclusion

The social significance of these findings is highly relevant, given the widespread use of artificial intelligence in our daily lives. Our results could change the way we perceive and interact with Large Language Models (LLMs). In particular, they could greatly expand our capacity to automate language-based tasks and offer more sophisticated interactions between humans and machines. The promising proficiency observed in LLMs like GPT4 suggest that these models could be employed to facilitate and enhance human-to-machine communication.

The examination of linguistic pragmatics has unequivocally established the effectiveness of Large Language Models (LLMs), particularly with respect to their practical application in simulating understanding of the human dialogues. This research has explored and compared various LLMs with human subjects within a theoretical framework established by Sperber and Wilson’s Relevance Theory and Gricean maxims (Sperber & Wilson, 1986, 1995; Grice, 1989). The results have indicated that LLMs, on average, performed significantly better than human subjects in simulating interpretation of dialogues for implicit and explicit meanings.

While the findings of this study are quite significant, indicating that the GPT model surpasses human performance in the aspect of linguistic pragmatics and demonstrates an understanding of communication and context of situations commonly encountered by humans, it is important to consider the broader picture. The assessment of these models isn't confined to linguistic pragmatics alone. Regardless of how impressive GPT model's performance has been in understanding and interpreting dialogues, it represents merely one facet of the wide array of cognitive abilities that constitute human reasoning and understanding. Therefore, further assessments encompassing other cognitive dimensions would be required to determine the overall capability of these models in truly replicating or even outperforming human cognition. The broader implications of these findings demand a comprehensive examination of AI capabilities to foster their potential benefits responsibly and ethically.

A noteworthy observation was the continuous improvement in performance and understanding in the progression of the LLM versions, with GPT4 emerging as the top performer. Its ability to simulate identification and correct interpretation of complex elements of the dialogues, particularly those flauting the conventional maxims, underlines its superior capacity in simulating linguistic pragmatics interpretation. This observation is profound considering that these models were not specifically trained for this task, which leads to the inevitable conclusion that these models have somehow managed to grasp pragmatic rules in their training phase.

In interpreting these findings, it is important to note that GPT2 and GPT3 were not designed or fine-tuned primarily to follow instructions. While their pretraining may have included some instruction-based tasks, they lack the explicit optimization for instruction-following that is present in GPT3.5, GPT4, and Bard. This difference in training approach may have contributed to the observed mismatch between our dialogue prompts and GPT2/GPT3's ability to accurately handle them.

All these findings contribute to an understanding of the growing capacity of LLMs to simulate understanding and work with human pragmatics. It strongly suggests that not only can these models enhance their utility in a variety of practical applications, but they can also serve as a tool for further research in the field of linguistic pragmatics.

Despite these remarkable findings, it is important to mention that work remains to be done. While the LLMs did outperform human subjects in this study, they are not infallible and did not always correctly simulate interpretation of the dialogues or simulate identification of the specific pragmatics' principles at play. LLMs' preference for verbose answers, as well as occasional errors in simulating identification of the party causing the communicative challenge, signify potential areas of improvement.

Modern research often advocates for using large corpora (e.g., from authentic web or spoken sources) precisely because these corpora reflect the complexities of naturally occurring discourse. We opted for made-up dialogues in this initial study to control for extraneous variables and keep the number of items manageable for human participants. However, we recognize that utilizing naturally occurring dialogue data could enhance ecological validity. Therefore, a logical follow-up would include confirming the present findings with genuine conversations drawn from broader corpora or recorded interactions. This expanded approach could further validate the performance of LLMs on spontaneous, unstructured discourse, reducing any potential influence of artificiality introduced by scripted scenarios.

Conclusively, this study has shed significant light on the capability of LLMs in linguistic pragmatics. Their progressive improvement, which has been demonstrated by the comparison across multiple task-oriented dialogues and has been further validated by experienced linguists, signifies the extent of their learning capacity. As these models continue to evolve, we can foresee their implication unfolding in a variety of practical applications. The precision in interpretation they have demonstrated reveals the exciting potential they hold for the future. With targeted improvements and continued exploration, Large Language Models can redefine the dynamics of language understanding, including its implicitities, complexities, and subtleties.

The advancements of general-purpose AI systems hold significant implications for the future design of humanoid robots. As multi-modal LLMs become increasingly adept at simulating human understanding and interpreting complex aspects of the physical and social world, they can be integrated into humanoid robots to facilitate more natural and effective interactions with humans. This integration could enable robots to simulate understanding not only of explicit instructions but also implicit meanings, social cues, and contextual nuances that are essential for human communication. Consequently, humanoid robots equipped with advanced AI could navigate complex social environments, respond appropriately to human emotions, and perform a broader array of tasks that require sophisticated language understanding and social interaction. This progression could broaden the utility of humanoid robots across various sectors such as healthcare, education, and customer service and brings us closer to human-robot coexistence, where robots can act as intuitive collaborators and companions in our daily lives.

The study's limitations include the non-random sampling of participants, differences in testing conditions, and the focus on English language only. Future research should aim to replicate the study with more diverse and larger participant samples, including varied cultural and linguistic backgrounds, to determine the generalizability of the findings.

Additionally, investigating the role of explicit instruction in pragmatics and its impact on performance could provide valuable insights. Exploring how factors like motivation, engagement, and testing environments influence task performance would further enhance the understanding of human and AI capabilities in linguistic pragmatics.

With the current trend of rapid advancements in AI, it is reasonable to expect even more capable iterations of LLMs in the future. Hence, continuous evaluation of these newer models, while expanding the scope of their testing to cover broader aspects of pragmatic communication, seems like a promising area of exploration. Endeavors to bridge the gap between LLMs and human-level understanding will undeniably revolutionize the realm of language comprehension and communication. This rapid progress of AI and its combination with various other technologies, apps, services, and products could have vast social, ethical, and policy implications that require continuous assessment and evaluation by scientists from various disciplines and regulators (Alipour et al., 2024; Bojic, 2022; Bojic, 2024; Bojic et al., 2024a, 2024b).

## Ethical Approval

All procedures performed in studies involving human participants were in accordance with the ethical standards of the institutional research committee and with the 1964 Helsinki Declaration and its later amendments or comparable ethical standards. The study was approved by the Ethics Committee of the Institute for Artificial Intelligence Research and Development of Serbia (August 1, 2023 / No. 01082023).

## Informed Consent:

Informed consent was obtained from all individual participants included in the study. Written consent was obtained from the first cohort of participants in person by Milan Cabarkapa, using an online form accessed on the computers in the room at the Faculty of Engineering Sciences, University of Kragujevac, on September 12, 2023. The second cohort provided written consent remotely via the same online form, administered by the Latenta Market Research Agency on January 29, 2025. The consent covered participation in the survey, as well as the use of data for further analysis and publication. Prior to participation, all participants received detailed information about the study's purpose, procedures, potential risks, and benefits. They were assured of the confidentiality and anonymity of their responses and informed that participation was voluntary. Consent was indicated by clicking a consent button.

## Competing Interests:

The authors declare that they have no competing interests.

## Author Contributions:

- Ljubiša Bojić: conceptualization, methodology, writing - original draft, investigation, qualitative analysis of the responses of LLM models and writing up the relevant part of the discussion section
- Predrag Kovačević: defining the analytical framework in terms of Gricean Maxims; writing up the linguistic portion of the theoretical background, designing and creating the stimuli (conversational snippets), qualitative analysis of the responses of LLM models and writing up the relevant part of the discussion section
- Milan Čabarkapa: supervision, survey administration, qualitative analysis of the responses of LLM models and writing up the relevant part of the discussion section

## Data Availability

The data that support the findings of this study are available from the Open Science Framework (OSF). The dataset is openly accessible at https://osf.io/96wrq/?view_only=21e230071e824c01be6c075e98d79503 and includes information on linguistic pragmatics, the Generative Pre-training Transformer family, and the Bard family. Researchers interested in exploring the dataset for further analysis or replication studies can access it directly via the aforementioned link.

**AI disclosure**

This study involved the use of several large language models (LLMs) to explore aspects of linguistic pragmatics. Specifically, OpenAI's GPT-2, GPT-3, GPT-3.5, and GPT-4, accessed via the OpenAI Playground with default settings, were utilized on August 30, 2023. Additionally, Google's Bard platform was employed to conduct the same set of experiments on August 31, 2023. For transparency, the responses generated during these sessions have been publicly released (OSF, 2023). To verify the consistency and reliability of the model outputs, the testing procedure was repeated on December 9, 2023, yielding comparable results. All model names, access dates, and details regarding the scope of use are provided herein to fully disclose the extent of AI involvement in the research process.